# Knock-Knock: Acoustic Object Recognition using Stacked Denoising Autoencoders

Shan Luo, Leqi Zhu, Kaspar Althoefer, Hongbin Liu

*Abstract*—This paper presents a successful application of deep learning for object recognition based on acoustic data. The shortcomings of previously employed approaches where handcrafted features describing the acoustic data are being used, include limiting the capability of the found representation to be widely applicable and facing the risk of capturing only insignificant characteristics for a task. In contrast, there is no need to define the feature representation format when using multilayer/deep learning architecture methods: features can be learned from raw sensor data without defining discriminative characteristics a-priori. In this paper, stacked denoising autoencoders are applied to train a deep learning model. Knocking each object in our test set 120 times with a marker pen to obtain the auditory data, thirty different objects were successfully classified in our experiment and each object was knocked 120 times by a marker pen to obtain the auditory data. By employing the proposed deep learning framework, a high accuracy of 91.50% was achieved. A traditional method using handcrafted features with a shallow classifier was taken as a benchmark and the attained recognition rate was only 58.22%. Interestingly, a recognition rate of 82.00% was achieved when using a shallow classifier with raw acoustic data as input. In addition, we could show that the time taken to classify one object using deep learning was far less (by a factor of more than 6) than utilizing the traditional method. It was also explored how different model parameters in our deep architecture affect the recognition performance.

## I. INTRODUCTION

Future intelligent robots are envisioned to be endowed with perceptive capabilities to see, touch and hear what is happening in the ambient world. This would enable robots to perform various tasks with object recognition being among the most common and significant ones. To perform this task, many types of sensors can be utilized and each kind of sensor offers a different view of objects. One of the richest and most widely used sensors is the camera as much information can be acquired from one single image. Because of this, vision has attracted considerable attention in object recognition by classifying the color [1], texture [2], [3], surface reflectance [4] and appearance [5], [6]. But vision is heavily dependent on the surrounding environments and would fail due to the variance of poses, illumination changes or occlusion by other objects. Another sensing modality with rich information content is the sense of touch. With the use of force/tactile sensors, the object properties can be revealed by accessing the information of hardness/softness [7], shape [8], thermal cues [9], surface texture [10] and surface friction properties [11]. However, tactile object recognition needs the direct contact with the

Shan Luo (corresponding author, mountluoupc@gmail.com), Leqi Zhu, Kaspar Althoefer, Hongbin Liu are with the Centre for Robotics Research, Department of Informatics, King's College London, WC2R 2LS, UK. Shan Luo is also with School of Civil Engineering and School of Computing, Univerisity of Leeds, LS2 9JT, UK. Kaspar Althoefer is also with School of Engineering and Materials Science, Queen Mary University of London, E1 4NS, London, UK.

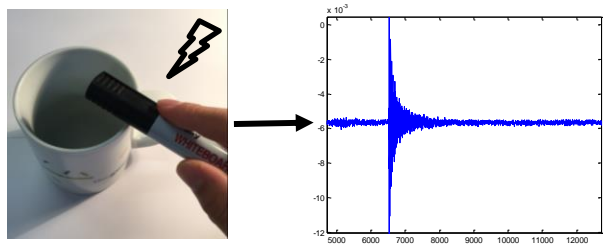

**Fig. 1.** An example of the acoustic data (right) collected by striking a marker pen on a cup (left).

objects and there is a risk of damage to objects when handling them, especially if they are fragile. Damage to the manipulator can occur if the handled object is impacting negatively on the robot. Acoustic data acquired by sensors like microphones represents an interesting alternative when trying to recognize objects. It allows the robot to work in a safer condition and slight taps on fragile objects can provide rich information without breaking them. Sound signals generated by striking an object can expose the intrinsic properties of objects such as elasticity and internal friction [12]. The elasticity of an object is directly related to the speed of sound waves in the object and therefore influences the frequency of the sound. The internal friction, or dampness, determines how the generated sounds decay over time [12] and provides shape-variant acoustic features for object classification [13].

To date, however, audition has been largely neglected when compared to vision and tactile sensing utilized in the area of object recognition. One of the most dominant factors is that the auditory data is more abstract compared to visual images and force/tactile data. In the traditional approach to acoustic based recognition, the task is achieved by using handcrafted features in the time [14] or frequency domain [15] usually using shallow classifiers. However, there are several drawbacks of such methods. Firstly, it is time consuming and laborious to extract the features. Secondly, it is difficult to design appropriate features for specific tasks. Thirdly, using features of pre-defined types can reduce the wide applicability of the method and may result in capturing characteristics of minor importance for a task. Fourthly, as for vision and tactile sensing, a multitude of acoustic features are present in sounds and these are organized in a hierarchical structure; therefore, the use of handcrafted features and shallow classifiers will lead to information loss. In order to learn abstract and hierarchical features automatically from raw sound data, we propose to apply deep learning for acoustic object recognition. The contributions of this paper can be summarized as:

(1) A novel method to recognize objects is provided by using deep learning based on acoustic data.

(2) The potential prospects of deep learning in acoustic based object recognition has been investigated and explored.

(3) Different model parameters in deep architecture that affect the recognition performance have been evaluated.

As depicted in Fig. 1, acoustic data was generated by striking the test object with a marker pen. As a result, an acoustic data vector of 1×N is acquired and fed as input to the deep network. The remainder of the paper is organized as follows. The related work on acoustic object recognition and deep learning based object recognition is reviewed in Section II. In Section III, the principle of Stacked Denoising Autoencoder (SDAE) and its application in audio based object recognition is introduced. And data collection procedure is presented in the following section. The results using deep learning are provided and the effect of different model parameters on the recognition performance is evaluated in Section V. It is compared with traditional sound recognition methods in Section IV. Finally, the paper is concluded in the last section.

## II. RELATED WORK

### A. Acoustic object recognition

When compared to object recognition based on visual and tactile information, less research has been conducted on acoustic-based object recognition. Nevertheless, audition is as important as the sense of touch and vision, and has distinct advantages especially in dark or hazardous environments. Burst et al. showed the feasibility to achieve object classification through sound information originating from striking objects [15]. The two most significant spikes in the power spectrum were extracted as features and their frequency coordinates were taken as inputs to a minimum-distance classifier. A reasonable classification accuracy of 94% was achieved, however, only five test objects were used. Following this work, in [12] and[13], both the spectral content and decay rate were exploited to achieve the perception of object materials from contact sounds. In [16], the recognition of objects was based on the distributions of characteristic resonances and decay rates. In these works, actions of a single type, e.g., striking, were taken. Different from this, Sinapov et al. introduced a series of exploration behaviors, including shaking, grasping, dropping, tapping and pushing, to obtain different sounds from household objects [17]. Self-organizing Maps (SOMs) were implemented for feature extraction and $k$-Nearest Neighbor and SVM classifiers were used for classification. The work was studied further in [18] by integrating acoustic data with proprioceptive torque feedback and a better recognition rate was achieved. In a recent study [19], spectral energies were taken as the base features and a general Fourier domain analysis borrowed from speech signal analysis was applied. In this work, the acoustic signals were generated by the interaction of a dexterous hexapod robot with the surfaces of different materials. In all the above works, complex preprocessing stages had to be applied to eliminate spurious signals and handcrafted features and shallow classifiers were utilized.

### B. Deep learning based object recognition

As deep learning can extract higher-level representation of sensory inputs, it has attracted increasing attention in object recognition and shown promising results in different applications. To recognize the objects present in natural images, Krizhevsky et al. took the raw image pixels as inputs and trained a large, deep convolutional neural network to classify the objects in the ImageNet data set and enhanced the state-of-the-art recognition rate from 73.9% to 84.7% [20]. A more recent work used deep convolutional neural networks to learn hierarchical features from RGB-D images for object recognition and pose estimation [21], showing superior results when compared to the traditional methods. In the view of tactile sensing, Schmitz et al. [22] applied deep learning in tactile object recognition and a dramatic performance improvement was observed in classifying 20 different objects compared to using traditional neural networks. However, to the best knowledge of the authors, research concerning acoustic object recognition by employing deep learning has not been attempted so far.

## III. METHODOLOGY

The deep learning framework for the acoustic object recognition chosen here consists of two phases. The first phase is to train each deep network layer as a denoising autoencoder (DAE) by unsupervised pre-training in a layerwiser manner. The second phase is to stack the latent representations learned in the first phase to form a deep network that is fine-tuned as a whole using back propagations. In this phase, only the encoding part of each autoencoder in the first phase is considered. In the deep network the nodes in the input layer are the raw acoustic data and in the output layer are object classes. Based on the learned deep network, the test objects can be classified. Both phases are introduced in detail as follows.

### A. Unsupervised pre-training

To begin with, the autoencoder, the base of the DAE, is first introduced. It first maps the input vector $x$ into latent representation $y$ through transformation:

$$y = s(Wx + b), \quad (1)$$

where $s$ is the activation function (hyperbolic tangent was used in our experiment), W and b are weight matrix and offset vector respectively. This is called encoding and the latent representation can be treated as a compressed representation of the input. After the encoding process, the latent representation is mapped back into a reconstruction $\hat{x}$ through a similar transformation:

$$\hat{x} = s(W'y + b'). \quad (2)$$

This process is called decoding and the output vector $\hat{x}$ can be interpreted as a prediction of the inputs $x$, given the latent representation. The output layer has equally many nodes as the input layer. An autoencoder tends to minimize the error in reconstructing its input $x$, i.e., to bring the output $\hat{x}$ close or make equal to input $x$. In the first autoencoder the input is the raw acoustic data and the obtained latent representation is fed as input to the second autoencoder layer. In this manner, the latent representations of the other autoencoders are acquired.

The denoising autoencoder is a stochastic variant of the classical autoencoder. As in traditional autoencoders, one aims to minimize the reconstruction loss between the input vector $x$ and its reconstruction from $y$. The difference is that y is acquired from the transformation of a corrupted input, as shown in Fig. 2. This process tries to undo the effect of a corruption process stochastically applied to the input of the auto-encoder whilst preserving the information encoded in the input. In other words, a DAE is trained to reconstruct a "repaired" input from the corrupted input and make the latent representations become more robust features [23]. As can be seen in Fig. 2, this can be done by adding random noise into original input $x$, i.e., setting some of the inputs to zero.

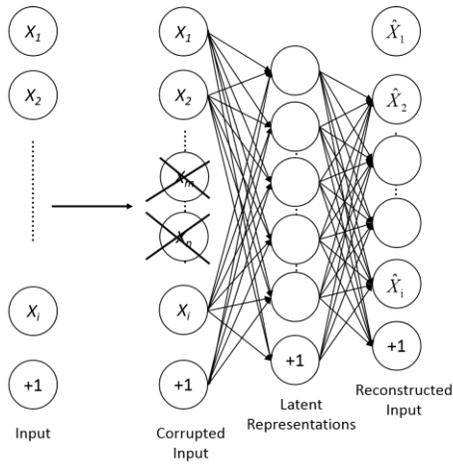

**Fig. 2.** An illustration of the denoising autoencoder structure. The input *x* is corrupted and the encoder is aimed to reconstruct *x* from the corrupted input.

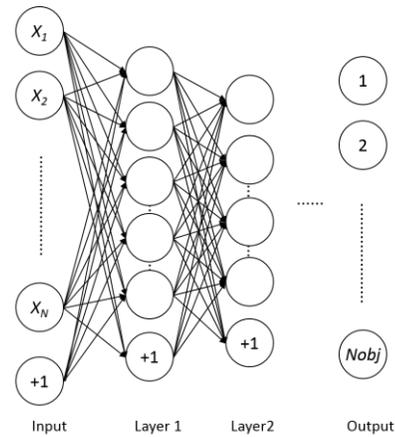

**Fig. 3.** An illustration of the deep structure. In this case, there are two hidden layers which are latent representations trained by denoising autoencoders separately. The input and output layers of the deep network are raw acoustic data and object classes respectively.

*B. Supervised fine-tuning*

Once all layers are pre-trained through DAEs, the deep network is constructed by stacking all the latent representation layers together as shown in Fig. 3. The input layer nodes are the raw acoustic data that are present in $1\times N$ vectors and the output nodes are object classes from 1 to *Nobj* that is the number of objects. The entire network is then fine-tuned in a supervised manner to minimize the error in predicting the object labels using back propagation. More details can be found in [23].

## IV. DATA COLLECTION

In our experiments, each data collection trial was carried out as follows. The test object was struck with a plastic marker pen and the generated impact sounds were recorded in Matlab via the mono channel of the microphone of a laptop. For each trial, the object was struck at different places. The sampling frequency was 8 kHz and the recording time for each trial was two seconds. As a result, a series of 16,000 data points in the range of [-1, 1] can be gained. To trim the redundant information in the data, only 500 points starting from the peak value of each sound signal, which can cover the whole knocking process for each trial, were taken as the input for deep learning model, therefore, $N=500$. The data collection process was conducted for 120 times for each object, with the first 100 times as the training phase and the remaining 20 times as the test phase. In total, thirty objects taken from the daily life were utilized in our experiments, as depicted in Fig. 4. It can be noticed that there are some objects of similar properties. For instance, object 1 and 2 are filled and unfilled bottles respectively; they have the same surface properties but have different density properties and, hence, different acoustic properties. It is very difficult to distinguish them by judging their visual appearance or employing static touch. For humans, it is very easy to utilize the impact sound generated by striking to differentiate one such object from the other. Therefore, the robot is also expected to possess such capacity by employing our deep learning framework. For the purpose of minimizing the influence of noise, all experiments were performed in an otherwise quiet room.

## V. EXPERIMENTAL RESULTS AND ANALYSIS

As presented in Section III, the SDAE model contains two main parts, the pre-training phase and the fine-tuning phase. To make the results more robust the classification process was conducted five times and the mean values of the results were taken. As a result of the structure of the collected data, in the deep network there were 500 nodes in the input layer and 30 nodes in the output layer. We investigated how different parameters in the deep learning model would affect the recognition performance including the number and layout of hidden layers, the number of hidden nodes, the number of iterations at pre-training and fine-tuning phases, the learning rates, and the impact of conducting the experiments with or without denoising. The parameters were optimized as shown in Table I according to the following three tips mentioned in [24]:

(1) Adjust one single parameter at one time;
(2) Scale consideration (e.g., learning rate of 0.1 and 0.2 may not differ much, but of 0.1 and 0.01 may have significant difference);
(3) Computational considerations.

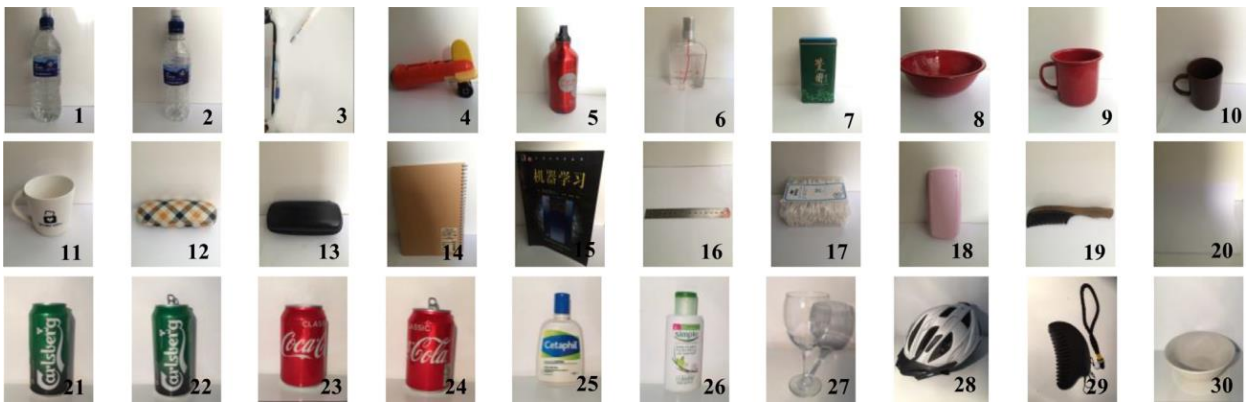

**Fig. 4.** Objects used for the experiments and they are labeled from 1 to 30 marked at the bottom right of the picture of each object. 1. Mineral water bottle full of water 2. Empty mineral water bottle 3. Table 4. Toy plane 5. Kettle 6. Perfume bottle 7. Tea box 8. Bowl_1 9. Cup_1 10. Cup_2 11. Cup_3 12. Glasses case_1 13. Glasses case_2 14. Book_1 15. Book_2 16. Ruler 17. Cotton box 18. Calculator 19. Wood comb 20. Paper 21. Unopened beer 22. Empty beer can 23. Unopened coke 24. Empty coke can 25. Lotion_1 26. Lotion_2 27. Wine glass 28. Helmet 29. Stone comb 30. Bowl_2.

The effect of the number of hidden layers was investigated and interestingly it was found that more layers could out always yield superior recognition performance, as shown in Fig. 5. As the number of hidden layers was increased from 1 to 3, the recognition performance was enhanced. The probable reason for it is that more latent representations can extract more abstract features from the raw data at this stage. However, as the number of hidden layers was further increased (i.e., beyond 3), the recognition performance deteriorated; this could be interpreted as the deep network being affected by the excessive description (overfitting). Therefore, a setting of three hidden layers was selected in our study. The other parameters were as shown in Table I.

There are three types of layouts of nodes in hidden layers: 1) increasing size, which is present in a shape of pyramid; 2) parallel size, in which all hidden layers have the same number of nodes; 3) decreasing size, which is present in an inverted pyramid shape. The effect of these three layout types was investigated and the results are shown in Table II. It should be noted that the first layer and last layer in all cases are for the sensory input (500 nodes) and object classes (30 nodes) respectively. It can be observed that the parallel structure performed the best. It means that the parallel layout of hidden layers is more suitable for acoustic object recognition. Hence, a parallel structure was also used in the subsequent tests.

TABLE I Optimized parameters

| Parameter | Value |
|---|---|
| Numbe of hidden layers | 3 |
| Layout of hidden layers | Parallel |
| Number of hidden nodes | 200 |
| Unsupervised pre-training epochs | 500 |
| Supervised fine-tuning epochs | 100 |
| Learning rates | 0.1 |
| Denoising or not | Denoising |

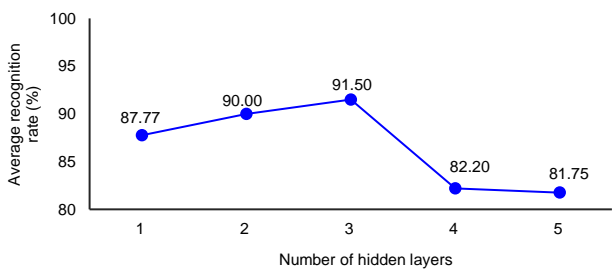

**Fig. 5.** Recognition rates with various number of hidden layers.

Compared to the layer structure, it was found that the variance of the number of nodes used in each layer had less effect on the recognition performance. It can be seen in Fig. 6 that the recognition rate has a slight change when the number of nodes in each hidden layer is increased from 200 to 400. However, the computational expense would increase as more nodes present in each layer. Therefore, 200 nodes were present in each hidden layer.

The number of pre-training epochs was proved to be important for the recognition performance. As shown in Fig. 7, the recognition rates were increased as the number of pre-training epochs was incremented. It means that the more pre-training epochs are taken the better representations can be extracted from the raw data. But the performance levels off when the number of pre-training epochs arrived at 500. Hence, 500 epochs were taken for the pre-training phase.

Nevertheless, the epochs at the fine-tuning stage were found to have less effect on the recognition performance. As illustrated in Fig. 8, there was only a small difference in the achieved recognition rates. Also taking the computational expense into consideration, 100 epochs were taken for the fine-tuning phase.

The learning rate of the pre-training phase was also considered as a significant parameter and its effect on the recognition performance is shown in Fig. 9. It could be divided into two phases: as the learning rate was incremented from 0.01 to 1 the recognition rate was increased whereas when the learning rate was greater than 0.1 the recognition performance deteriorated dramatically. Hence, the learning rate was set to 1 in our study.

TABLE II Recognition rates with different layer structure

| Layout | Recognition rate |
|---|---|
| 500-100-200-300-30 | 4.50% (overfitting) |
| 500-100-100-100-30 | 91.50% |
| 500-300-200-100-30 | 73.83% |

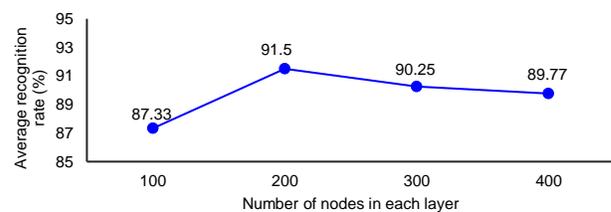

**Fig. 6.** Recognition rates with various number of nodes in each hidden layer.

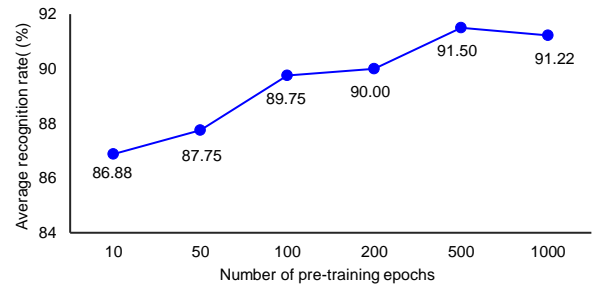

**Fig. 7.** Recognition rates with different pre-training epochs.

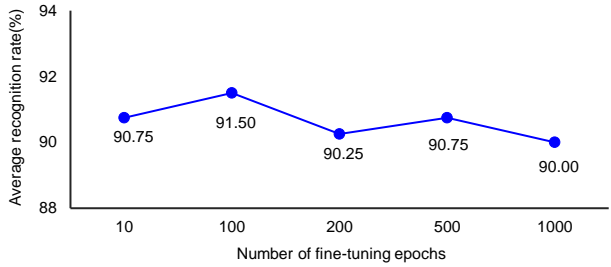

**Fig. 8.** Recognition rates with different fine-tuning epochs.

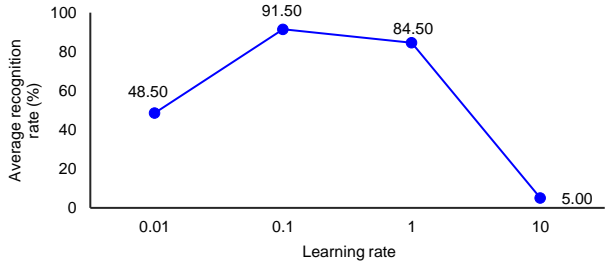

**Fig. 9.** Recognition rates with different learning rates.

In addition, we investigated the impact of the process of denoising in the pre-training phase on the recognition performance. It was observed that the framework with denoising outperformed the one without denoising, with an improvement of 3.75% in the recognition rate. It indicates that the inclusion of denoising makes the learned deep learning model more robust.

Based on the above discussions, the optimized parameters were obtained as listed in Table I. As a result, an overall classification accuracy of 91.50% was achieved; a confusion matrix is shown in Fig. 10. It proves that our proposed deep learning framework can exploit the latent feature representations of the raw acoustic data and the objects can be recognized accurately. It can be observed that some objects that are difficult to distinguish by using vision or tactile sensing, e.g., filled and unfilled bottles (objects 1 and 2), can be classified successfully. On the other hand, only a few of the objects are assigned to wrong labels, e.g., some observations of the paper (object 20) are wrongly concluded to be from the kettle (object 5).

## VI. COMPARISON WITH TRADITIONAL METHODS

As has been mentioned in Section I, traditional acoustic recognition is achieved by using handcrafted features with shallow classifiers. This approach is employed in this section and compared with our proposed deep learning framework. In this paper, we used the Mel-Frequency Cepstral Coefficients (MFCCs) [25] and its first and second differentials as features. It can well describe the nonlinear characteristics of the human ear frequency and it is popular in traditional sound processing.

A Hanning window with a 32-ms fixed frame rate was first applied to the acquired acoustic signal to perform a Fourier transform. After that, we used $12^{th}$-order MFCC together with its first and second temporal derivatives as features. As a result, each feature was a 36 dimensional vector. The features were then implemented with a SVM classifier using LibSVM [26]. The best recognition rate using this method was only 58.22%. In addition, complex preprocessing process had to be applied and the feature extraction process needed to be designed elaborately.

In an additional experiment, we utilized the raw acoustic data as the input of the SVM classifier; the recognition results using different methods are listed in Table III. It was surprising that a high recognition rate of 91.50% was achieved - much better than it was the case using MFCC features. A probable reason is that the original structure of the acoustic data appears to be more distinctive for shallow classifiers. But the recognition performance of the SVM-based classifier applied to raw data was still inferior to that of our proposed deep learning framework.

We also compared the time for classifying test objects - an important aspect when considering real time applications. All algorithms were implemented in MATLAB and executed on a laptop with a 1.4Ghz Intel Core i5 processor and 4GB DDR3-1600 RAM. The time taken to classify test objects (excluding time for training the deep model) using our proposed deep learning framework was found to be much shorter. For classifying the thirty objects (20 trials for each), the minimum time taken was 0.03189s and the maximum time taken was 0.05450 s using the deep learning framework. In comparison, by using SVM with raw data, the minimum time taken was 0.36884s. Hence, the classification using deep learning framework was 6.77 faster than the latter, as shown in Table IV. This inspiring result indicates that when a larger dataset is investigated, the strength of deep learning becomes obvious.

|    | 1 | 2 | 3 | 4 | 5 | 6 | 7 | 8 | 9 | 10 | 11 | 12 | 13 | 14 | 15 | 16 | 17 | 18 | 19 | 20 |
|----|---|---|---|---|---|---|---|---|---|----|----|----|----|----|----|----|----|----|----|----|
| 1  | 1.00 | 0.00 | 0.00 | 0.00 | 0.00 | 0.00 | 0.00 | 0.00 | 0.00 | 0.00 | 0.00 | 0.00 | 0.00 | 0.00 | 0.00 | 0.00 | 0.00 | 0.00 | 0.00 | 0.00 |
| 2  | 0.00 | 0.95 | 0.00 | 0.00 | 0.05 | 0.00 | 0.00 | 0.00 | 0.00 | 0.00 | 0.00 | 0.00 | 0.00 | 0.00 | 0.00 | 0.00 | 0.00 | 0.00 | 0.00 | 0.00 |
| 3  | 0.00 | 0.00 | 1.00 | 0.00 | 0.00 | 0.00 | 0.00 | 0.00 | 0.00 | 0.00 | 0.00 | 0.00 | 0.00 | 0.00 | 0.00 | 0.00 | 0.00 | 0.00 | 0.00 | 0.00 |
| 4  | 0.00 | 0.00 | 0.00 | 0.95 | 0.00 | 0.00 | 0.00 | 0.00 | 0.00 | 0.00 | 0.00 | 0.00 | 0.00 | 0.00 | 0.00 | 0.00 | 0.05 | 0.00 | 0.00 | 0.00 |
| 5  | 0.00 | 0.00 | 0.00 | 0.00 | 0.85 | 0.00 | 0.00 | 0.00 | 0.00 | 0.00 | 0.00 | 0.00 | 0.00 | 0.00 | 0.00 | 0.10 | 0.00 | 0.00 | 0.05 | 0.00 |
| 6  | 0.00 | 0.00 | 0.00 | 0.00 | 0.00 | 1.00 | 0.00 | 0.00 | 0.00 | 0.00 | 0.00 | 0.00 | 0.00 | 0.00 | 0.00 | 0.00 | 0.00 | 0.00 | 0.00 | 0.00 |
| 7  | 0.00 | 0.00 | 0.00 | 0.00 | 0.00 | 0.00 | 1.00 | 0.00 | 0.00 | 0.00 | 0.00 | 0.00 | 0.00 | 0.00 | 0.00 | 0.00 | 0.00 | 0.00 | 0.00 | 0.00 |
| 8  | 0.00 | 0.00 | 0.00 | 0.00 | 0.00 | 0.00 | 0.00 | 0.95 | 0.00 | 0.00 | 0.00 | 0.00 | 0.00 | 0.00 | 0.00 | 0.00 | 0.00 | 0.05 | 0.00 | 0.00 |
| 9  | 0.00 | 0.00 | 0.00 | 0.00 | 0.00 | 0.00 | 0.00 | 0.00 | 1.00 | 0.00 | 0.00 | 0.00 | 0.00 | 0.00 | 0.00 | 0.00 | 0.00 | 0.00 | 0.00 | 0.00 |
| 10 | 0.00 | 0.00 | 0.00 | 0.00 | 0.00 | 0.00 | 0.00 | 0.00 | 0.00 | 1.00 | 0.00 | 0.00 | 0.00 | 0.00 | 0.00 | 0.00 | 0.00 | 0.00 | 0.00 | 0.00 |
| 11 | 0.00 | 0.00 | 0.00 | 0.00 | 0.00 | 0.00 | 0.00 | 0.05 | 0.00 | 0.00 | 0.95 | 0.00 | 0.00 | 0.00 | 0.00 | 0.00 | 0.00 | 0.00 | 0.00 | 0.00 |
| 12 | 0.00 | 0.00 | 0.00 | 0.00 | 0.00 | 0.00 | 0.00 | 0.00 | 0.00 | 0.00 | 0.00 | 0.80 | 0.10 | 0.00 | 0.00 | 0.00 | 0.00 | 0.10 | 0.00 | 0.00 |
| 13 | 0.00 | 0.00 | 0.00 | 0.00 | 0.00 | 0.00 | 0.00 | 0.00 | 0.00 | 0.00 | 0.00 | 0.00 | 1.00 | 0.00 | 0.00 | 0.00 | 0.00 | 0.00 | 0.00 | 0.00 |
| 14 | 0.00 | 0.00 | 0.00 | 0.00 | 0.00 | 0.00 | 0.00 | 0.00 | 0.00 | 0.00 | 0.00 | 0.00 | 0.00 | 1.00 | 0.00 | 0.00 | 0.00 | 0.00 | 0.00 | 0.00 |
| 15 | 0.00 | 0.00 | 0.00 | 0.00 | 0.00 | 0.00 | 0.00 | 0.00 | 0.00 | 0.00 | 0.00 | 0.00 | 0.00 | 0.00 | 1.00 | 0.00 | 0.00 | 0.00 | 0.00 | 0.00 |
| 16 | 0.00 | 0.00 | 0.00 | 0.00 | 0.00 | 0.00 | 0.00 | 0.00 | 0.00 | 0.00 | 0.05 | 0.00 | 0.00 | 0.00 | 0.00 | 0.95 | 0.00 | 0.00 | 0.00 | 0.00 |
| 17 | 0.00 | 0.00 | 0.00 | 0.05 | 0.00 | 0.00 | 0.00 | 0.00 | 0.00 | 0.00 | 0.00 | 0.00 | 0.00 | 0.00 | 0.00 | 0.00 | 0.95 | 0.00 | 0.00 | 0.00 |
| 18 | 0.00 | 0.00 | 0.00 | 0.00 | 0.00 | 0.00 | 0.00 | 0.00 | 0.00 | 0.00 | 0.00 | 0.00 | 0.00 | 0.00 | 0.00 | 0.00 | 0.00 | 0.95 | 0.05 | 0.00 |
| 19 | 0.00 | 0.00 | 0.00 | 0.00 | 0.00 | 0.00 | 0.00 | 0.00 | 0.00 | 0.00 | 0.00 | 0.00 | 0.00 | 0.00 | 0.00 | 0.00 | 0.00 | 0.00 | 1.00 | 0.00 |
| 20 | 0.00 | 0.00 | 0.00 | 0.00 | 0.25 | 0.00 | 0.00 | 0.00 | 0.00 | 0.05 | 0.00 | 0.00 | 0.00 | 0.00 | 0.00 | 0.00 | 0.00 | 0.00 | 0.00 | 0.70 |

**Fig. 10.** Confusion matrix of object recognition. The ground truths of the object labels are listed in the vertical axis while estimations are listed in the horizontal axis. The object labels are consistent with the ones in Fig. 4.

TABLE III Recognition rates with different methods

| Method | Recognition rate |
| --- | --- |
| Deep learning | 91.50% |
| SVM with MFCC features | 58.22% |
| SVM with raw data | 82.00% |

Table IV Time taken for classifying test objects

| Method | Time/s |
| --- | --- |
| Deep learning | 0.05450 (longest) |
| SVM with raw data | 0.36884 (shortest) |

## VII. CONCLUSIONS AND FUTURE WORK

This paper proposes a deep learning based method for the acoustic object recognition. Based on Stacked Denoising Autoencoders through both unsupervised pre-training and supervised fine-tuning, a multi-layer nonlinear mapping structure of deep network is trained to automatically extract high-level and more abstract features from the original acoustic data. It is proved that this deep learning based method can achieve better recognition performance compared to the traditional method using handcrafted features with a shallow classifier. It can be seen that the recognition rate increased 33.28% through deep learning without the need to employ a complex feature extraction process. It is also worth mentioning that the test time is dramatically faster using deep learning than using traditional method (our approach is by a factor of 6.77 faster). In addition, various parameters in the deep learning network were investigated. In our experiment, there was no clear evidence to show that more layers would lead to better recognition performance. The parameters, including layout of hidden layers, number of hidden nodes, number of iterations at pre-training and fine-tuning phases, learning rates, and the setting of with or without denoising, were also studied.

There are several branches for future research. Compared to the large image recognition dataset like ImageNet, our dataset might not be big enough, thus it is planned to increase our dataset in the future work, not only by increasing the number of trials of striking, but also by enlarging the number of test objects. Deeper neural networks have achieved good results in the image classification task, e.g., the deep residue nets achieved 3.57% error on the ImageNet test set. Thus, it is also worth trying deeper neural nets in the acoustic object recognition [27]. In the current work most of the objects we chose are with relatively solid surface. However, for the soft or deformable objects such as clothes and teddy bear, it might become challenging because striking soft objects can only generate tiny impact sound. Hence, another extension of our work could be combining the auditory cues with other sensing information such as texture through tactile sensing.